\documentclass[nonacm,sigconf]{acmart}

\AtBeginDocument{%
  \providecommand\BibTeX{{%
    \normalfont B\kern-0.5em{\scshape i\kern-0.25em b}\kern-0.8em\TeX}}}

\copyrightyear{2020}
\acmYear{2020}
\setcopyright{rightsretained}
\acmConference[HSCC '20]{23rd ACM International Conference on Hybrid Systems: Computation and Control}{April 22--24, 2020}{Sydney, NSW, Australia}
\acmBooktitle{23rd ACM International Conference on Hybrid Systems: Computation and Control (HSCC '20), April 22--24, 2020, Sydney, NSW, Australia}
\acmDOI{10.1145/3365365.3382220}
\acmISBN{978-1-4503-7018-9/20/04}

\newcommand{\todo}[1]{}
\renewcommand{\todo}[1]{{\color{red} TODO: {#1}}}

\newcommand{\tool}{\texttt{dtControl}}

\newcommand{\scots}{\texttt{SCOTS}}
\newcommand{\uppaal}{\textsc{Uppaal Stratego}}
\newcommand{\stratego}{\textsc{Uppaal Stratego}}

\DeclareMathOperator*{\argmax}{arg\,max}
\DeclareMathOperator*{\argmin}{arg\,min}
\newcommand{\eqdef}{\mathrel{\mathop:}=}

\newcommand{\tree}{\mathsf{T}}
\newcommand{\pred}{\rho}
\newcommand{\lab}{\lambda}
\newcommand{\Reals}{\mathbb{R}}

\newcommand{\actions}{\mathcal{U}}
\newcommand{\action}{u}
\newcommand{\states}{X}
\newcommand{\st}{\vec{x}}

\newcommand{\entropy}{\texttt{entr}}
\newcommand{\predicates}{\texttt{PREDS}} \usepackage{algpseudocode,algorithm}
\usepackage{amsmath}

\usepackage{tikz}
\usetikzlibrary{positioning,arrows,plotmarks}
\usepackage{tkz-euclide}

\usepackage{booktabs}
\usepackage{multirow}

\usepackage{subcaption}
\usepackage{float}
\usepackage{listings}

\usepackage{enumitem}

\begin{document}

\title[\tool: Decision Tree Learning Algorithms for Controller Representation]{\tool: Decision Tree Learning Algorithms\\ for Controller Representation}

\author{Pranav Ashok}%
\author{Mathias Jackermeier}
\author{Pushpak Jagtap}
\author{Jan K\v{r}et\'{i}nsk\'{y}}
\author{Maximilian Weininger}
\affiliation{%
  \institution{Technical University of Munich}
  \streetaddress{Boltzmannstr. 3}
  \city{Munich}
  \country{Germany}
  \postcode{85748}
}
\author{Majid Zamani}
\affiliation{%
  \institution{University of Colorado Boulder}
  \streetaddress{1111 Engineering Drive}
  \city{Boulder}
  \country{USA}
  \postcode{80309}
}
\affiliation{%
  \institution{Ludwig Maximilian University of Munich}
  \city{Munich}
  \country{Germany}
}

\renewcommand{\shortauthors}{Ashok et al.}
\newtheorem{remark}{Remark}

\begin{abstract}
Decision tree learning is a popular classification technique most commonly used in machine learning applications. 
Recent work has shown that decision trees can be used to represent provably-correct controllers concisely. 
Compared to representations using lookup tables or binary decision diagrams, decision trees are smaller and more explainable. 
We present \tool, an easily extensible tool for representing memoryless controllers as decision trees.
We give a comprehensive evaluation of various decision tree learning algorithms applied to 10 case studies arising out of correct-by-construction controller synthesis.
These algorithms include two new techniques, one for using arbitrary linear binary classifiers in the decision tree learning, and one novel approach for determinizing controllers during the decision tree construction. 
In particular the latter turns out to be extremely efficient, yielding decision trees with a single-digit number of decision nodes on 5 of the case studies.
\end{abstract}


\maketitle

\section{Introduction}
Formal synthesis of controllers enforcing complex specifications on cyber-physical systems has gained significant attention in the last few years. 
This is mainly due to the need for obtaining formally verified control strategies rendering some complex tasks; these are usually represented using temporal logic specifications or (in)finite strings over automata. 
There are several techniques and tools available that provide automated, correct-by-construction, controller synthesis for cyber-physical systems by utilizing symbolic models (a.k.a.\ finite abstractions) \cite{tabuada2009verification,belta2017formal}, in which the uncountable continuous states and inputs are aggregated to finite symbolic states and inputs via quantization (a.k.a.\ discretization). 
The so-called symbolic controllers are then computed by utilizing algorithmic machinery from computer science and then mapped back for use in the original systems. The state-of-the-art tools to synthesize such controllers are, e.g., \scots~\cite{SCOTS:RunggerZ16}, {\tt pFaces}~\cite{khaled2019pfaces}, {\tt QUEST}~\cite{jagtap2017quest}, {\tt Pessoa}~\cite{mazo2010pessoa}, {\tt CoSyMA}~\cite{mouelhi2013cosyma}, or {\uppaal}~\cite{stratego}.
These tools give a huge list of state-action pairs (a.k.a. lookup tables) representing~ controllers. 
Storing these symbolic controllers in the memory is a major problem because they usually need to run on embedded devices with limited memory.
However, if we do not store the controllers as lookup tables, but take advantage of decision trees (DT) \cite{mitchellML}, which exploit their hidden structure
to represent them in a more compact way, we can mitigate this problem.
As shown in \cite{sos:AKL+19}, DTs can be orders of magnitude smaller than lookup tables.
Such a concise representation opens the door for better readability, understandability, and explainability of the controllers, while reducing memory requirements and preserving correctness guarantees.
Moreover, human-understandable controllers may also provide insight into the models themselves, thus aiding their validation, as we illustrate in the example below.

Our setting is inherently different from the usual use of DT in machine learning; 
there, in order to generalize well, DTs typically do not fit the training data exactly;
in contrast, in this work, DTs have to exactly represent the given controllers in order to preserve their correctness guarantee.
Therefore, our requirements on DTs differ: beside the size and the explainability, it is also the \emph{perfect fitting}.
Consequently, it is necessary to thoroughly re-evaluate current DT-learning algorithms and possibly also modify them.

A basic technique used to represent controllers more concisely is to \emph{determinize} them, i.e. to make them not (maximally) permissive but only retain a single action for each state.
To this end, one can use, for instance, the action with the minimum norm from a reference input, when least energy consuming controllers are preferred \cite{philipp}, or the previously applied action (if possible), when lazy controllers are preferred \cite{mazo2010pessoa,mouelhi2013cosyma}.
Such a size reduction by determinization can be applied as \emph{pre-processing before} learning the DT representation of the controller, typically yielding also a smaller DT.
Alternatively, one can apply other kinds of reduction by determinization as \emph{post-processing after} constructing the DT.
For instance, in ``safe pruning'' of \cite{sos:AKL+19}, the DT constructed for the maximally permissive controller is modified as follows.
The leaves of the tree are merged in a bottom-up fashion, thereby reducing the size and partially determinizing it.
In contrast, here we introduce a novel approach for determinizing the controllers \emph{during} the construction of the DT, with advantages to both pre-processing and post-processing methods.
Firstly, since the choice of the action for each state greatly affects the size and structure of the DT, it is advantageous to guide the choice by the concrete, already built part of the DT, compared to a-priori choices made by pre-processing approaches.
Secondly, while the post-processing approaches have to construct a large tree first, our new technique constructs an already reduced tree, avoiding the intermediate large one, thus making it more scalable. %

\paragraph{Motivating Example}

\begin{figure}[t]
\begin{tikzpicture}

\node (A) at (0, 0) {$T_{room2} \leq 20.625$};
\node[below left=0.5cm and 0cm of A,] (B) {$T_{room5} \leq 20.625$};
\node[below right=0.5cm and 0cm of A] (C) {$T_{room5} \leq 20.625$};
\node[below left=0.35cm and 0cm of B,] (D) {$(1,1)$};
\node[below right=0.35cm and 0cm of B] (E) {$(1,0)$};
\node[below left=0.35cm and 0cm of C] (F) {$(0,1)$};
\node[below right=0.35cm and 0cm of C] (G) {$(0,0)$};

\draw[-] (A) -- node[above left] {true} (B);
\draw[-,dashed] (A) -- node[above right] {false} (C);
\draw[-] (B) -- node[above left] {} (D);
\draw[-,dashed] (B) -- node[above right] {} (E);
\draw[-] (C) -- node[above left] {} (F);
\draw[-,dashed] (C) -- node[above right] {} (G);
\end{tikzpicture}
\caption{Decision tree for the temperature~controller}
\label{fig:10rooms}
\end{figure}
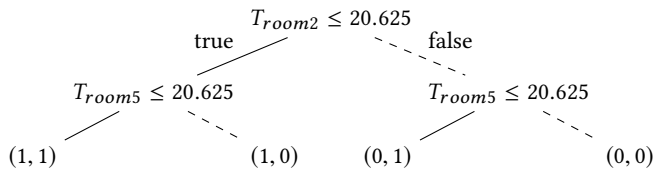

Consider a temperature control system running in a building with 10 rooms with the heater installed only in 2 rooms as described in \cite{jagtap2017quest}.
The permissive controller maintaining the temperatures of all the rooms within a certain range obtained using \scots\ is a lookup table with 52,488 state-action pairs. 
By naively determinizing, we get a lookup table with 26,244 symbolic states (i.e. domain of the controller) and their respective actions.
The standard DT-learning, e.g. \cite{CART:BreimanFOS84}, applied to these two lookup tables yields DT with 8,648 and with 2,703 decision nodes, respectively.
While this is an improvement, it is far from being explainable. 
With the help of our novel determinization strategy presented in Section \ref{sec:nondet}, we are able to obtain the decision tree with only 3 (!) decision nodes, see Figure \ref{fig:10rooms}. 
Apart from obtaining a compact and easily implementable controller representation while preserving correctness guarantees, the result is so small that it is immediately explainable and, moreover, allows us to improve on the implementation: 
one can readily see that we only need to install temperature sensors in two rooms instead of all 10 rooms, which will help users to reduce the system deployment cost as well as the required bandwidth to transfer the state information to the controller. Only 4 symbols (leaves of the tree) need to be transferred~to~realize~the~controller. 
We also obtain a controller with very few nodes for the cruise-control model of \cite{cruise:LarsenMT15}.
From such a clear representation one immediately notices that the controller makes the car decelerate when the car in front of it is far away. 
This counter-intuitive behaviour has thus revealed a bug in the model, which did not actually describe the intended behaviour~of~the~system. 
\smallskip

The contribution of this paper can be summarized as follows:
\begin{itemize}
    \item We present \tool, an open-source tool to convert formally verified controllers to decision trees preserving their correctness guarantees.
    \tool\ has a simple input format and already supports automated conversion for controllers generated by two state-of-the-art tools -- \uppaal~\cite{stratego} and \scots~\cite{SCOTS:RunggerZ16}.
    It supports several output formats, most importantly the graphical output as DOT files, useful for further analysis and visual presentation, and the C source code, useful for closed-loop simulation or for loading onto embedded devices.
    \item We introduce a new technique for using arbitrary binary classifiers in the DTs and a novel approach for determinizing controllers during the DT learning.~%
    Our approach is tuned towards obtaining extremely small, explainable DTs.
    In 5 out of 8 case studies where it is applicable (the original controllers are non-deterministic), it produces trees with single-digit numbers of decision nodes. 
    \item We present a comprehensive evaluation of 8 DT-learning algorithms on 10 case studies.
\end{itemize}

\paragraph{Related Work}
DTs \cite[Chapter 3]{mitchellML} are a well-known class of data structures, particularly known for their interpretability, used mostly by machine learning practitioners in classification 
or regression 
tasks. %
Our work is based on well-known algorithms for decision tree learning, namely CART~\cite{CART:BreimanFOS84}, C4.5~\cite{C4.5:Quinlan93} and OC1~\cite{OC1}.

There has been previous work on combining decision trees with classifiers, namely 
Perceptrons~\cite{perceptrontrees}, Logistic Regression models~\cite{logistictrees}, piece-wise functions~\cite{neider} or Support-Vector Machines~\cite{svmtree,strategyrep:ABC+19}.
We generalize those approaches by allowing for arbitrary binary classifiers to be used in our trees. 
Additionally, those methods are either restricted to only use two labels, which is not applicable for controllers with more than two possible actions, or they only allow linear classifiers in leaf nodes \cite{strategyrep:ABC+19, neider}.
In contrast, our approach is applicable with an arbitrary number of actions and also leverages the power of linear classifiers in inner nodes.

An alternative to DTs are binary decision diagrams (BDD) \cite{BDD_Bryant86}. As seen in \cite{sos:AKL+19, BrazdilCKT18,cav15jan}, BDDs have several disadvantages: firstly, they do not retain the inherent flavour of decisions of strategies as maps from states to actions due to their bit-level representation and, hence, are hardly explainable.
Secondly, they are notoriously hard to minimize \cite{BrazdilCKT18}, also because finding the best variable ordering is NP-complete~\cite{BDD_Bryant86}.
BDDs only allow binary classification, so the actions have to be joined with the state space to represent a controller. The recent result in \cite{zapreev} discusses various heuristic-based determinization algorithms for BDDs representing controllers; however, they still suffer from those disadvantages we mentioned for BDDs.
Algebraic decision diagrams (ADD)~\cite{ADDs} are an extension of BDDs that allow to have more than two labels, i.e. associate every action to a leaf node. However, they still suffer from the same drawbacks as BDDs.
In~\cite{girard2013low} ADDs are used for controller representation; however, no concrete algorithm is provided. 

The formal methods community has made use of decision trees 
to represent controllers and counterexamples arising out of model checking Markov decision processes, stochastic games and LTL synthesis \cite{cav15jan,strategyrep:ABC+19,sos:AKL+19,BrazdilCKT18}.
DTs have also been used to represent learnt policies from reinforcement learning \cite{pyeatt2001decision}. However, in contrast to our paper, \cite{pyeatt2001decision} does not preserve safety guarantees, only considers axis-aligned splits and does not consider non-determinism. \cite{DBLP:journals/corr/abs-1810-04240} suggests the possibility of using regression trees for representing policies, whereas we consider classification trees.
\section{Tool}
\tool\ is an easy-to-use open-source tool for post-processing memoryless symbolic controllers into various compact and more interpretable representations. 
We report the input and output formats as well as the algorithms that are currently supported. Note that the tool can easily be extended with new formats and algorithms. 
\tool\ is distributed as an easy-to-install \texttt{pip} package\footnote{\texttt{pip} is a standard package-management system used to install and manage software packages written in Python. See \url{https://pypi.org/project/dtcontrol/}.} along with a user and developer manual\footnote{Available at \url{https://dtcontrol.readthedocs.io/en/latest/}}.

\subsubsection*{Dependencies}
\tool\ works with Python version 3.6.7 or higher. The core of the tool which runs the learning algorithms require \texttt{numpy}, \texttt{pandas} and \texttt{scikit-learn}~\cite{scikit-learn}.
Optionally, \tool\ may also require the C-based oblique decision tree tool \texttt{OC1}~\cite{OC1}.

\subsubsection*{Input formats}
\tool\ currently accepts controllers in three formats: (i) a raw comma-separated values (CSV) format with each row consisting of a vector of state variables concatenated with a vector of input variables;
(ii) a sparse matrix format used by \scots; and (iii) the raw strategy produced by \uppaal. More details about the various formats are described in the user manual.

\subsubsection*{Algorithms}
\tool\ offers a range of parameters to adjust the DT learning algorithm, which are described in Section \ref{sec:concrete}. 

\subsubsection*{Output formats}
\tool\ outputs the decision tree in the DOT graph representation language (for visual presentation of the tree), as well as \texttt{C} 
code that can be directly used for implementation; see Appendix \ref{app:output} for the DOT and \texttt{C} output that \tool\ produces for the DT in Figure \ref{fig:10rooms}.
Additionally, \tool\ reports statistics for every constructed tree, namely size, the minimum number of bits required to represent symbols in obtained controller, and the construction time. 
\section{Preliminaries - Decision tree learning} \label{sec:dtLearning}

A decision tree (DT) over the domain $\states$ with the set of labels $\actions$ is a tuple $(\tree,\lab,\pred)$, 
where $\tree$ is a finite full binary tree (every node has exactly 0 or 2 children),
$\lab$ assigns to every leaf node (node with 0 children) a label $\action \in \actions$ and 
$\pred$ assigns to every inner node (node with 2 children, also called decision node) of the tree a predicate, which is a boolean function $\states \mapsto \{0,1\}$.

The semantics of a DT is as follows: 
given a state $\st$, there is a unique \emph{decision path} through the tree $\tree$ starting from the root node (the only node with no parent) to a leaf node $\ell$.
This means that the label for state $\st$ is $\lab(\ell)$.
The decision path is defined by starting at the root node, and then for each decision node $n$ evaluating the predicate on the state, i.e. computing $\pred(n)(\st)$, and picking the left child if the predicate is true and the right child otherwise.%

For example, consider the DT in Figure \ref{fig:10rooms}: 
$\tree$ has 7 nodes, 3 of which are decision nodes (including the root node) and 4 of which are leaf nodes.
A state of the system is a vector of 10 temperatures, e.g. $\st = (20.1,20.2,20.3,20.4,20.5,20.6,20.7,20.8,20.9,21.0).$
To find the decision for this state, we first evaluate the predicate in the root node.
Since the temperature in the second room is smaller than 20.625, the predicate is true and we go to the left child. 
We evaluate the next predicate in the same fashion and arrive at the leaf node labelled $(1,1)$, which gives us a safe control input, in this case to turn on both heaters.

All DT learning algorithms implemented in \tool\ follow the same underlying structure:
given a finite set $C\subseteq X\times\mathcal{U}$ of feature-label pairs, it returns a DT that represents $C$ precisely; this means that for every $(\st,\action) \in C$, the leaf node of the decision path for $\st$ has the label $\action$.
In the setting of this paper, $C$ is a controller, features are states and labels are actions\footnote{We use the term actions instead of control inputs, to avoid confusion because of the fact that the control inputs are the outputs of a DT.}.

To learn the DT, the algorithm tries to minimize the entropy of $C$, denoted $\entropy(C)$, by splitting it according to a predicate.
Formally, for some $C \subseteq \{(\st,\action) \mid \st \in \states, \action \in \actions\}$, 
$$\entropy(C) \eqdef - \sum_{\action \in \actions} p_{\action} \log(p_{\action}),$$ 
where $p_{\action} \eqdef \frac{\lvert \{(\st, \action) \in C\} \rvert} {\lvert C \rvert}$ is the empirical probability of label $\action$ being in $C$; notation $|\cdot|$ denotes the cardinality of a set.
The underlying algorithm works recursively as follows:
\begin{itemize}
    \item \textbf{Base case:}
        If $\entropy(C) = 0$, i.e. all pairs $(\st, \action) \in C$ have the same label $\action$, then return the following DT: the tree $\tree$ has only a single node $r$, with $\lab(r) = y$, and $\pred$ has no domain in this case, as there are no decision nodes.
    \item \textbf{Recursive case:}
        If $\entropy(C) \neq 0$, $C$ needs to be split; for that, we use some predicate $P \in \predicates$ which splits $C$, where the set $\predicates$ to be picked here is a parameter of the algorithm that is discussed in Section \ref{sec:preds}.
        We pick the predicate that minimizes the entropy after the split, i.e.,
        $$\argmin_{P \in \predicates} \entropy(\{(\vec{x}, u) \in C \mid P(\vec{x})\})
        + \entropy(\{(\vec{x}, u) \in C \mid \neg P(\vec{x})\}).$$
        Intuitively, the best predicate is the one which is able to split $C$ into two parts which are as homogeneous as possible. 
        Given the best predicate, we recursively call the algorithm on the subsets resulting from the split, getting two DTs $(\tree_t,\lab_t,\pred_t)$ and $(\tree_f,\lab_f,\pred_f)$; the indices $t$ and $f$ indicate whether the predicate was true or false, respectively.
        Then we return the following DT: the tree $\tree$ has the root node $r$, with the left child being the root of $\tree_t$ and the right child the root of $\tree_f$.
        $\lab$ uses $\lab_t$ for leaves of the left sub-tree and $\lab_f$ for the right sub-tree.
        $\pred$ is defined similarly on the inner nodes of the left and right sub-trees, with the addition that $\pred(r) = P$, i.e. the predicate of the root of $\tree$ is the predicate we used for the split.
\end{itemize}

The symbolic controllers designed by \scots\ and \uppaal\ are generated by correct-by-construction synthesis procedures. 
In order to use these controllers for original systems (i.e. with infinite continuous states and inputs), we need to refine the controllers. 
For more details on refinement procedures, we kindly refer the interested reader to~ \cite{reissig2016feedback,tabuada2009verification,euler}. 

\tool\ preserves the correctness guarantees by representing the symbolic controllers precisely, i.e. iterating until the entropy in all leaf nodes is 0.
In the case of determinization, \tool\ represents one of the deterministic sub-controllers
precisely, which is chosen on-the-fly during the construction.

\section{Methods}\label{sec:concrete}
There are two parameters of \tool:
the set of predicates to consider ($\predicates$) and the way in which non-determinism is handled.
For each of these, \tool\ implements existing ideas and introduces new ones.
Here, we only report the high-level ideas; for a more detailed description, refer to the user or developer manual.

\subsection{Predicates}\label{sec:preds}
\subsubsection{Existing idea: Axis-aligned splits}
 In the standard algorithms, e.g \cite{CART:BreimanFOS84,C4.5:Quinlan93}, only \emph{axis-aligned splits} are considered; i.e. predicates that can only have the form $x_i \sim b$, where $x_i$ is one of the state variables, $b \in \Reals$, and $\sim \,\in \{\leq,\geq\}$.
In our setting, the set of possible predicates is greatly restricted due to discretization (quantization). 
The number of splits to be evaluated for each variable $x_i$ is equal to the number of discrete values of $x_i$.
        
\subsubsection{Existing idea: Oblique splits} 
Beside the standard axis-aligned splits, \tool\ also supports predicates of the form $\vec{w}^{T} \vec{x} \leq b$, where $\vec{w}, \vec{x} \in \Reals^{n}, b \in \Reals$. 
These \emph{oblique predicates} \cite{OC1} incorporate information from multiple state variables in a single split and thus have the potential to greatly simplify the induced decision tree~\cite{strategyrep:ABC+19}. 
However, due to combinatorial explosion, it is too costly to simply enumerate all possible oblique predicates even in the discretized space, due to which different heuristics are employed~\cite{OC1}. 
In this regard, \tool\ supports the usage of predicates obtained using (an adapted version of) the OC1 algorithm~\cite{OC1}.

\subsubsection{New technique: Using binary machine-learnt classifiers}

It is possible to find non-axis-aligned predicates splitting the controller by using classification techniques from machine learning. 
As our main goal is for the resulting tree to be explainable, we want to avoid complex predicates, and thus we restrict the classifiers we consider in two ways:
(i) we only consider linear classifiers, and %
(ii) we restrict to binary classifiers, so that the resulting tree is binary.

We use these binary linear classifiers in a way that is similar to the classical one-vs-the-rest classification, e.g. \cite[Chapter 4]{Bishop}:
For each action $\action$, we train a classifier $LC_\action$ that tries to separate the states with that action from the rest. 
We then pick that classifier whose predicate minimizes the entropy, i.e.
\[
\boxed{
LC \eqdef \argmin_{\action \in \actions} 
\begin{aligned}
\entropy(\{(\st, \action) \in C &\mid LC_\action(\st) = 1\})\\
&+\\
\entropy(\{(\st, \action) \in C &\mid LC_\action(\st) = 0\}).
\end{aligned}
}
\]

We considered various linear classification techniques including Logistic Regression~\cite[Chapter 4]{Bishop}, linear Support Vector Machines (SVM)~\cite[Chapter 7]{Bishop}, Perceptrons~\cite[Chapter 5]{Bishop}, and Naive Bayes~\cite{NaiveBayes}.
However, the latter two yielded significantly larger DTs in all of our experiments, so \tool\ does not offer these algorithms to the end-user.

\smallskip

In summary, \tool\ currently supports four possibilities for the set $\predicates$:
axis-aligned predicates, the modified oblique split heuristic from~\cite{OC1} and oblique splits obtained either via logistic regression or linear SVM classifiers. Due to the modular structure of the code, it is easy to extend the existing approaches or add new methods, as described in our developer manual.

\subsection{Non-determinism}\label{sec:nondet}

In the general algorithm described in Section \ref{sec:dtLearning}, for the sake of simplicity, we restricted our procedure to controllers that deterministically choose a single control input. In case of non-deterministic (also called permissive) controllers, the tuples in the controller $C$ have the form $(\st,\action)$, where $\action$ is now a set $\{\action_1,\action_2,\dots,\action_m\}$ of admissible control inputs. One approach to handle non-determinism is to simply assign a \emph{unique label} to each set, and hence reduce the setting to the case where for every state there is only a single label. This means that the DT algorithm can be used in exactly the same way as described in Section~\ref{sec:dtLearning}. This method retains all information that was initially present in the given controller.

The disadvantage of handling non-determinism like this is that the number of unique classes may be as large as $2^{|\actions|}$.
In order to avoid this blow-up and optimize memory, one can decide to determinize the controller.
If we have some knowledge about which value of a control input is optimal, e.g. from domain knowledge or since it was computed by an \emph{optimization algorithm} as in \stratego~\cite{stratego}, this information can be used, eliminating the non-deterministic choice.
Otherwise, one can use a standard determinization approaches, e.g. picking the value with the \emph{minimum norm}. 
The tree can then simply be constructed from the determinized labels. 
Additionally, we propose the following alternative to these determinization approaches.

\paragraph{Novel determinization approach: Maximal frequencies}
Our new determinization technique MaxFreq aims to minimize the size of the resulting DT. The underlying general idea is simple: 
if many of the data points share the same label, a DT learning algorithm should group them together under the common label.
This idea naturally gives a determinizing strategy when applied in our context.

Consider a set $C$ of pairs of state and sets of actions.
The goal is to identify for each state a single action which can be assigned to it. 
Let $f$ be the function for action frequency, which maps actions to their number of occurrences in $C$. 
Then, for each state $\st$ such that $(\st, \{\action_1,\action_2,\ldots,\action_m\}) \in C$, we re-assign to $\st$ the single label $\action'$ which appears with the highest frequency. 
Formally, our determinization procedure produces for each state $\st$, an action $\action'(\st)$, where
\[
\boxed{\forall (\st, \{\action_1, \ldots, \action_m\}) \in C.\, \action'(\st) = \argmax_{\action \in \{\action_1, \ldots, \action_m\}} f(\action).}
\]

Once we have determinized $C$, we can use any method presented in Section~\ref{sec:preds} to find a predicate for the current node. 
After the set is split, the procedure is recursively applied to both child nodes, recomputing the action frequency each time.

\smallskip

In summary, \tool\ offers 3 different possibilities to handle non-determinism:
unique labels retaining the information, determinizing upfront by picking the action with the minimal norm, and using the novel heuristic MaxFreq. 
\section{Experiments}\label{sec:exp}

\begin{table*}[t]
    \caption{Result of running the various methods on 10 different case studies. The `Lookup table' column gives the size of the domain of the original controller. For all other columns, the number of decision paths in the constructed tree is indicated. The case studies are grouped together by the number of control inputs and methods based on whether they preserve non-determinism.
    $\infty$ indicates that the computation did not finish within 3 hours; n/a indicates that the approach is not applicable (we cannot determinize, as the model is already deterministic).}
    \label{tab:experiments}
    \begin{tabular}{lrrrrrrrrr}
    	\toprule
    	                                                                           &        \multicolumn{5}{c}{Most permissive controller}        &          \multicolumn{4}{c}{Determinized controller}           \\
    	\cmidrule(r{4pt}){2-6} \cmidrule(l{4pt}){7-10}
    	Case Study & Lookup table &    CART & LinSVM &           LogReg & OC1 &      MaxFreq &       MaxFreqLC & MinNorm & MinNormLC \\ \midrule
    	\multicolumn{3}{l}{\textbf{Single-input non-deterministic}}                                         &        &                  &     &              &                 &         &           \\
    	cartpole \cite{jagtap2018software}                                                                  &          271 &     127 &    126 &              100 &  92 &   \textbf{6} &      7 &      56 &        39 \\
    	2D Thermal \cite{girard2013low}                                                                     &       40,311 &      14 &     14 &               8 &  12 &   5 &      \textbf{4} &       8 &         \textbf{4} \\
    	helicopter \cite{jagtap2018software}                                                                &      280,539 &   3,174 &  2,895 &            1,877 &   $\infty$ & \textbf{115} &             134 &     677 &       526 \\
    	cruise \cite{cruise:LarsenMT15}                                                                     &      295,615 &     494 &    543 &              392 & 374 &   \textbf{2} &      \textbf{2} &     282 &       197 \\
    	dcdc \cite{SCOTS:RunggerZ16}                                                                        &      593,089 &     136 &    140 &              70 &  90 &   \textbf{5} &      \textbf{5} &      11 &       11 \\
    	\multicolumn{3}{l}{\textbf{Multi-input non-deterministic}}                                          &        &                  &     &              &                 &         &           \\
    	10D Thermal \cite{jagtap2017quest}                                                                  &       26,244 &   8,649 &     67 &              74 &   2,263 &   \textbf{4} &              10 &   2,704 &        28 \\
    	truck\_trailer\cite{khaled2019pfaces}                                                               &    1,386,211 & 169,195 & $\infty$ &        $\infty$ & $\infty$ &       21,598 & \textbf{12,611} &  95,417 &    30,888 \\
    	traffic\cite{swikir2019compositional}                                                               &   16,639,662 &   6,287 & $\infty$ &            4,477 & $\infty$ & 98 &         \textbf{80} &     690 &    $\infty$ \\
    	\multicolumn{3}{l}{\textbf{Multi-input deterministic}}                                              &        &                  &     &              &                 &         &           \\
    	vehicle \cite{SCOTS:RunggerZ16}                                                                     &       48,018 &   6,619 &  6,592 &   5,195 & \textbf{4,886} &            n/a &           n/a &       n/a &     n/a \\
    	aircraft \cite{rungger2015state}                                                                    &    2,135,056 & 456,929 & $\infty$ & \textbf{407,523} &   $\infty$ &            n/a &           n/a &       n/a &     n/a \\ \bottomrule
    \end{tabular}
\end{table*} 
All experiments were conducted on a server running on an Intel Xeon W-2123 processor with a clock speed of 3.60GHz and 64 GB RAM. 
We ran the unique-label approach with all 4 possible predicate classes (see Section \ref{sec:preds}): axis-aligned predicates (CART)~\cite{CART:BreimanFOS84}, oblique predicates with linear support-vector machines (LinSVM), logistic regression (LogReg), and the heuristic from~\cite{OC1}, called OC1.
Note that all these resulting trees represent the maximally permissive controller for the finite abstraction.
Additionally, on all the non-deterministic models we ran our novel determinization approach (see Section \ref{sec:nondet}) with axis-aligned predicates (MaxFreq), and with oblique predicates (MaxFreqLC where LC stands for linear classifier).
For the results in Table \ref{tab:experiments}, we used logistic regression as linear classifier, because it reliably performed well.
As a competitor for our determinization approach we use a-priori determinization with the minimum norm, again both with axis-aligned predicates (MinNorm) and with logistic regression for linear predicates (MinNormLC).
Additionally, we compare to the random a-priori determinization, to get an impression for possible cases where MinNorm would not be a natural choice but no better is given.
However, since the results are always worse, we only report the numbers in Appendix \ref{app:exp}.
Since some of the algorithms rely on randomization, we ran all experiments thrice and report the median.

We run the discussed algorithms on ten case studies, five of which are marked as multi-input,  containing control inputs which are multi-dimensional, i.e. $u = (u_1, \dots, u_m)$. All our algorithms work by giving each multi-dimensional control input a single action label, and then working on these labels as in the case of single-dimensional control inputs.

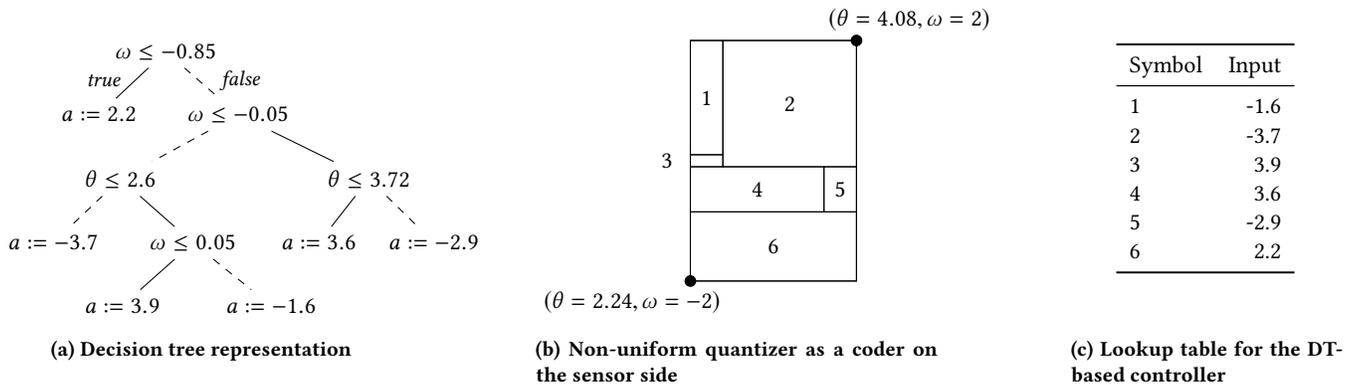
\begin{figure*}[ht]
	\centering
	\subcaptionbox{Decision tree representation\label{fig:cartpole-maxfreq}}[0.3\linewidth]{
\begin{tikzpicture}[x=1cm,y=0.45cm]

\node (A) at (0, 0) {$\omega \leq -0.85$};
\node[below left=0.9 and -0.5 of A,] (B) {$a:=2.2$};
\node[below right=0.9 and -0.6 of A] (C) {$\omega \leq -0.05$};
\node[below right=0.9 and 0.3 of C] (D) {$\theta \leq 3.72$};
\node[below left=0.9 and 0.2 of C] (E) {$\theta \leq 2.6$};
\node[below left=0.9 and -0.4 of E] (K) {$a:=-3.7$};
\node[below left=0.9 and -0.6 of D] (F) {$a:=3.6$};
\node[below right=0.9 and -0.5 of D] (G) {$a:=-2.9$};
\node[below right=0.9 and -0.3 of E] (H) {$\omega \leq 0.05$};	
\node[below left=0.9 and -0.35 of H] (I) {$a:=3.9$};
\node[below right=0.9 and -0.35 of H] (J) {$a:=-1.6$};

\draw[-] (A) -- node[pos=0.4,xshift=-12] {\small\emph{true}} (B);
\draw[-,dashed] (A) -- node[pos=0.4,xshift=16] {\small\emph{false}} (C);
\draw[-] (C) -- (D);
\draw[-,dashed] (C) -- (E);
\draw[-] (D) -- (F);
\draw[-,dashed] (D) -- (G);
\draw[-] (E) -- (H);
\draw[-] (H) -- (I);
\draw[-,dashed] (H) -- (J);
\draw[-,dashed] (E) -- (K);
\end{tikzpicture} 	}
	\hfill
	\subcaptionbox{Non-uniform quantizer as a coder on the sensor side\label{fig:cartpole-partitioning}}[0.3\linewidth]{
\begin{tikzpicture}[scale=1,x=1.2cm,y=0.8cm]
	\draw [draw=black] (2.24,-2) rectangle (4.08,2);
	\draw [draw=black] (2.24,-2) rectangle (4.08,-0.85) node[pos=.5] {6}; %
	\draw [draw=black] (2.24,-0.85) rectangle (3.72,-0.1) node[pos=.5] {4}; %
	\draw [draw=black] (3.72,-0.85) rectangle (4.08,-0.1) node[pos=0.5] {5}; %
	\draw [draw=black] (2.24,-0.1) rectangle (2.6,0.1) node[pos=.5,xshift=-15] {3}; %
	\draw [draw=black] (2.24,0.1) rectangle (2.6,2) node[pos=.5] {1}; %
	\draw [draw=black] (2.6,-0.1) rectangle (4.08,2) node[pos=.5] {2}; %
	\draw[mark=*,mark size=2pt] plot coordinates {(2.24,-2)} node[xshift=-22pt,yshift=-8pt] {$(\theta=2.24,\omega=-2)$};
	\draw[mark=*,mark size=2pt] plot coordinates {(4.08,2)} node[xshift=20pt,yshift=8pt] {$(\theta=4.08, \omega=2)$};
\end{tikzpicture} 	}
	\hfill
	\subcaptionbox{Lookup table for the DT-based controller \label{tab:cartpole-decoder}
}[0.2\linewidth]{
		\begin{tabular}{lr}
			\toprule
			Symbol & Input \\
			\midrule
			1      & -1.6          \\
			2      & -3.7          \\
			3      & 3.9           \\
			4      & 3.6           \\
			5      & -2.9          \\
			6      & 2.2          
\\
			\bottomrule
		\end{tabular}
		\vspace{2em}
	}
	\caption{End-to-end usage of DT-based controller: First, a DT representation is synthesized with the help of \tool\ (the result of running MaxFreq on cartpole is shown here). Then a non-uniform quantizer is implemented at the sensor side, which for each decision path (i.e. a region in the state-space), sends a state symbol to the controller. At the controller, this symbol gives actual control input. In this case, the information needs to be sent over the sensor-controller channel is $\lceil\log_2(6)\rceil = 3$ bits per time unit. The theoretical lower bound on the data rate in this example is $1$ bit per time unit to achieve invariance \cite{tomar2017invariance}.}\label{fig:cartpole}
\end{figure*}

In order to compare the sizes of the representations of the controllers fairly, we provide two different ways.
Firstly, the straight-forward way is to compare the number of nodes used in the DT and the number of rows in the lookup table, which we do in Table \ref{tab:app} in Appendix \ref{app:exp}.
However, a practically more relevant comparison should reflect the number of state symbols needed to capture the behaviour of the controller; these can also be directly related to memory requirements.
To this end, in Table \ref{tab:experiments} for DTs we report the number of decision paths, as these induce a partitioning of the state space into symbolic states.
For more information on this and an example, see Figure \ref{fig:cartpole} and the discussion in Section \ref{sec:discu}.

Beside comparing DTs to the lookup tables, we also compare them to BDDs.
However, BDDs do not directly correspond to the state symbols.
Hence we refrain from the state-symbols comparison and do not report BDD sizes in Table \ref{tab:experiments}, but only in Appendix \ref{app:exp}.
There, we compare the number of nodes in the BDDs to the number of nodes (not decision paths) generated by our DT algorithms.
The BDDs were generated using \scots~for all models but the two from \uppaal, cruise and 2D Thermal; for these two, we used the \texttt{dd} and \texttt{autoref} Python libraries.
The BDDs were minimized as much as possible by calling reordering heuristics until convergence.
The results show that the DT algorithms which determinize or which do not use oblique predicates are more scalable, as they were able to compute the result for all case studies, while BDDs timed out on dcdc and traffic.
Depending on the case study, BDDs are usually in the same order of magnitude as CART, sometimes better, sometimes worse. 
On the one hand, on 10D Thermal and truck\_trailer, BDDs have an order of magnitude less nodes, but on the other hand CART is able to produce results for dcdc and traffic.
Compared to MaxFreq, there is the exception of truck\_trailer, where the best BDD has a quarter of the size; on all other models, MaxFreq is at least one order of magnitude better.

\section{Discussion}\label{sec:discu}
Table \ref{tab:experiments} shows that DTs are always better than lookup tables. In the case of DTs exactly representing the most permissive controller, our linear-classifier-based algorithm, LogReg, generally performs better than the standard DT learning algorithm CART. 
An inspection of the trees showed that oblique splits indeed aid in this reduction. 
In order to save memory, however, our determinizing algorithms may be used. Here, MaxFreq and its linear classifier variant, MaxFreqLC, easily outperform all other discussed algorithms, returning trees which can be drawn on a single sheet of paper in most of our case studies! The controller produced by MaxFreq for the case study cartpole is depicted in Figure \ref{fig:cartpole-maxfreq}. 

Apart from the compact representation of the controllers and efficient determinization, \tool\ makes controllers more understandable. This helps to do some analysis for the systems and corresponding controllers. A few analyses were mentioned for the temperature control example in the introduction. 
Another application is that \tool\ learns how to efficiently partition the state space. In general, the tools synthesizing symbolic controllers use uniform partitioning, i.e. a uniform quantizer is used to discretize the state set. Therefore, they need a large number of symbols to represent the state set. \tool~ aggregates state symbols where the same control input is admissible to reduce the number of symbols required. In other words, \tool~ provides a scheme to design non-uniform quantizers (i.e., state encoders with non-uniform partitioning of state-set), illustrated in Figure \ref{fig:cartpole-partitioning}.

The entries in Table \ref{tab:experiments} correspond to the necessary number of state symbols. 
For instance, consider the cartpole example in Table \ref{tab:experiments}. The controller obtained using \scots~ requires $271$ symbols to represent the domain of the controller, which implies that one needs to send 9 bits per time unit over the sensor-controller channel to achieve invariance. 
After processing the controller using \tool~ with MaxFreq, we only need 6 symbols to represent the controller, corresponding to only 3 bits information. 
One can directly relate this idea of constructing efficient static coders to the notion of invariance feedback entropy introduced in \cite{tomar2017invariance}. This notion characterizes
the necessary state information required by any coder-controller to enforce the invariance condition in the closed loop. For example, in the case of cartpole, the theoretical lower-bound on average bit rate for any static coder-controller to achieve invariance is 1 (obtained through the invariance feedback entropy \cite{tomar2017invariance}), which is not far from 3, computed using \tool. 

In summary, one can utilize the results provided in this paper for constructing efficient coder-controllers for invariance properties which is an active topic in the domain of information-based control~\cite{NairFagniniZampieriEvans07}. 

\section{Conclusion}

We presented \tool, an open-source, easily extensible tool for post-processing controllers synthesized by various tools such as \scots\ and \uppaal\ into small, efficient and interpretable representations. 
The tool allows for a comparison between various representations in terms of size and performance and also allows us to export the controller both as a graphic and as a code. 
We also presented a new determinization technique, MaxFreq, which easily converts non-deterministic controllers into extremely small deterministic decision trees.
Further algorithms for controller representation were thoroughly evaluated and made accessible to the end-user.
We believe these small representations will not only allow us to save memory but also help us in understanding and validating the model.
As for future work, \tool\ can be extended with
\begin{itemize}
    \item further input and output formats, to also support tools such as {\tt pFaces}\cite{khaled2019pfaces} and {\tt QUEST}\cite{jagtap2017quest};
    \item different predicates: this can be other, possibly even non-linear or non-binary, machine-learning classifiers  or richer algebraic predicates utilizing domain knowledge;
    \item other \emph{impurity measures} instead of entropy, which decide the predicate used for the split
\end{itemize}

\begin{acks}
This work was supported in part by the H2020 ERC Starting Grant \emph{AutoCPS} (grant agreement no 804639), the German Research Foundation (DFG) through the grants ZA 873/1-1 and  KR 4890/2-1 \emph{Statistical Unbounded Verification}, and the TUM International Graduate School of Science and Engineering (IGSSE) grant 10.06 \emph{PARSEC}.
\end{acks}

\bibliographystyle{IEEEtran}
\bibliography{ref}

\begin{thebibliography}{10}
\providecommand{\url}[1]{#1}
\csname url@samestyle\endcsname
\providecommand{\newblock}{\relax}
\providecommand{\bibinfo}[2]{#2}
\providecommand{\BIBentrySTDinterwordspacing}{\spaceskip=0pt\relax}
\providecommand{\BIBentryALTinterwordstretchfactor}{4}
\providecommand{\BIBentryALTinterwordspacing}{\spaceskip=\fontdimen2\font plus
\BIBentryALTinterwordstretchfactor\fontdimen3\font minus
  \fontdimen4\font\relax}
\providecommand{\BIBforeignlanguage}[2]{{%
\expandafter\ifx\csname l@#1\endcsname\relax
\typeout{** WARNING: IEEEtran.bst: No hyphenation pattern has been}%
\typeout{** loaded for the language `#1'. Using the pattern for}%
\typeout{** the default language instead.}%
\else
\language=\csname l@#1\endcsname
\fi
#2}}
\providecommand{\BIBdecl}{\relax}
\BIBdecl

\bibitem{tabuada2009verification}
P.~Tabuada, \emph{Verification and control of hybrid systems: a symbolic
  approach}.\hskip 1em plus 0.5em minus 0.4em\relax Springer Science \&
  Business Media, 2009.

\bibitem{belta2017formal}
C.~Belta, B.~Yordanov, and E.~A. Gol, \emph{Formal methods for discrete-time
  dynamical systems}.\hskip 1em plus 0.5em minus 0.4em\relax Springer, 2017,
  vol.~89.

\bibitem{SCOTS:RunggerZ16}
M.~Rungger and Z.~M, ``{SCOTS:} {A} tool for the synthesis of symbolic
  controllers,'' in \emph{{HSCC}}.\hskip 1em plus 0.5em minus 0.4em\relax
  {ACM}, 2016, pp. 99--104.

\bibitem{khaled2019pfaces}
M.~Khaled and M.~Zamani, ``{pFaces}: an acceleration ecosystem for symbolic
  control,'' in \emph{Proceedings of the 22nd ACM International Conference on
  Hybrid Systems: Computation and Control}.\hskip 1em plus 0.5em minus
  0.4em\relax ACM, 2019, pp. 252--257.

\bibitem{jagtap2017quest}
P.~Jagtap and M.~Zamani, ``{QUEST}: A tool for state-space quantization-free
  synthesis of symbolic controllers,'' in \emph{International Conference on
  Quantitative Evaluation of Systems}.\hskip 1em plus 0.5em minus 0.4em\relax
  Springer, 2017, pp. 309--313.

\bibitem{mazo2010pessoa}
M.~Mazo, A.~Davitian, and P.~Tabuada, ``Pessoa: A tool for embedded controller
  synthesis,'' in \emph{International Conference on Computer Aided
  Verification}.\hskip 1em plus 0.5em minus 0.4em\relax Springer, 2010, pp.
  566--569.

\bibitem{mouelhi2013cosyma}
S.~Mouelhi, A.~Girard, and G.~G{\"o}ssler, ``{CoSyMA}: a tool for controller
  synthesis using multi-scale abstractions,'' in \emph{Proceedings of the 16th
  international conference on Hybrid systems: computation and control}.\hskip
  1em plus 0.5em minus 0.4em\relax ACM, 2013, pp. 83--88.

\bibitem{stratego}
A.~David, P.~G. Jensen, K.~G. Larsen, M.~Mikucionis, and J.~H. Taankvist,
  ``Uppaal stratego,'' in \emph{{TACAS}}, ser. Lecture Notes in Computer
  Science, vol. 9035.\hskip 1em plus 0.5em minus 0.4em\relax Springer, 2015,
  pp. 206--211.

\bibitem{mitchellML}
T.~M. Mitchell, \emph{Machine learning}, ser. McGraw Hill series in computer
  science.\hskip 1em plus 0.5em minus 0.4em\relax McGraw-Hill, 1997.

\bibitem{sos:AKL+19}
P.~Ashok, J.~K{\v{r}}et{\'i}nsk{\'y}, K.~G. Larsen, A.~Le~Co{\"e}nt, J.~H.
  Taankvist, and M.~Weininger, ``{SOS}: Safe, optimal and small strategies for
  hybrid markov decision processes,'' in \emph{{QEST} {(1)}}, D.~Parker and
  V.~Wolf, Eds.\hskip 1em plus 0.5em minus 0.4em\relax Springer, 2019, pp.
  147--164.

\bibitem{philipp}
P.~J. Meyer, M.~Rungger, M.~Luttenberger, J.~Esparza, and M.~Zamani,
  ``Quantitative implementation strategies for safety controllers,''
  \emph{arXiv preprint:1712.05278}, 2017.

\bibitem{CART:BreimanFOS84}
L.~Breiman, J.~H. Friedman, R.~A. Olshen, and C.~J. Stone, \emph{Classification
  and Regression Trees}.\hskip 1em plus 0.5em minus 0.4em\relax Wadsworth,
  1984.

\bibitem{cruise:LarsenMT15}
K.~G. Larsen, M.~Mikucionis, and J.~H. Taankvist, ``Safe and optimal adaptive
  cruise control,'' in \emph{Correct System Design}, ser. Lecture Notes in
  Computer Science, vol. 9360.\hskip 1em plus 0.5em minus 0.4em\relax Springer,
  2015, pp. 260--277.

\bibitem{C4.5:Quinlan93}
J.~R. Quinlan, \emph{{C4.5:} Programs for Machine Learning}.\hskip 1em plus
  0.5em minus 0.4em\relax Morgan Kaufmann, 1993.

\bibitem{OC1}
S.~K. Murthy, S.~Kasif, S.~Salzberg, and R.~Beigel, ``{OC1:} {A} randomized
  induction of oblique decision trees,'' in \emph{{AAAI}}.\hskip 1em plus 0.5em
  minus 0.4em\relax {AAAI} Press / The {MIT} Press, 1993, pp. 322--327.

\bibitem{perceptrontrees}
P.~E. Utgoff, ``Perceptron trees: {A} case study in hybrid concept
  representations,'' in \emph{{AAAI}}.\hskip 1em plus 0.5em minus 0.4em\relax
  {AAAI} Press / The {MIT} Press, 1988, pp. 601--606.

\bibitem{logistictrees}
N.~Landwehr, M.~A. Hall, and E.~Frank, ``Logistic model trees,'' in
  \emph{{ECML}}, ser. Lecture Notes in Computer Science, vol. 2837.\hskip 1em
  plus 0.5em minus 0.4em\relax Springer, 2003, pp. 241--252.

\bibitem{neider}
D.~Neider, S.~Saha, and P.~Madhusudan, ``Synthesizing piece-wise functions by
  learning classifiers,'' in \emph{International Conference on Tools and
  Algorithms for the Construction and Analysis of Systems}.\hskip 1em plus
  0.5em minus 0.4em\relax Springer, 2016, pp. 186--203.

\bibitem{svmtree}
I.~T. Christou and S.~Efremidis, ``An evolving oblique decision tree ensemble
  architecture for continuous learning applications,'' in \emph{{AIAI}}, ser.
  {IFIP}, vol. 247.\hskip 1em plus 0.5em minus 0.4em\relax Springer, 2007, pp.
  3--11.

\bibitem{strategyrep:ABC+19}
P.~Ashok, T.~Br{\'a}zdil, K.~Chatterjee, J.~K{\v{r}}et{\'i}nsk{\'y}, C.~H.
  Lampert, and V.~Toman, ``Strategy representation by decision trees with
  linear classifiers,'' in \emph{{QEST} {(1)}}.\hskip 1em plus 0.5em minus
  0.4em\relax Springer, 2019, pp. 109--128.

\bibitem{BDD_Bryant86}
R.~E. Bryant, ``Graph-based algorithms for boolean function manipulation,''
  \emph{IEEE Transactions on Computers}, vol. 100, no.~8, pp. 677--691, 1986.

\bibitem{BrazdilCKT18}
T.~Br{\'{a}}zdil, K.~Chatterjee, J.~Kret{\'{\i}}nsk{\'{y}}, and V.~Toman,
  ``Strategy representation by decision trees in reactive synthesis,'' in
  \emph{{TACAS} {(1)}}, ser. Lecture Notes in Computer Science, vol.
  10805.\hskip 1em plus 0.5em minus 0.4em\relax Springer, 2018, pp. 385--407.

\bibitem{cav15jan}
T.~Br{\'{a}}zdil, K.~Chatterjee, M.~Chmelik, A.~Fellner, and
  J.~Kret{\'{\i}}nsk{\'{y}}, ``Counterexample explanation by learning small
  strategies in markov decision processes,'' in \emph{{CAV} {(1)}}, ser.
  Lecture Notes in Computer Science, vol. 9206.\hskip 1em plus 0.5em minus
  0.4em\relax Springer, 2015, pp. 158--177.

\bibitem{zapreev}
I.~S. Zapreev, C.~Verdier, and M.~Mazo, ``Optimal symbolic controllers
  determinization for {BDD} storage,'' in \emph{ADHS}, 2018.

\bibitem{ADDs}
R.~I. Bahar, E.~A. Frohm, C.~M. Gaona, G.~D. Hachtel, E.~Macii, A.~Pardo, and
  F.~Somenzi, ``Algebraic decision diagrams and their applications,''
  \emph{Formal Methods in System Design}, vol.~10, no. 2/3, pp. 171--206, 1997.

\bibitem{girard2013low}
A.~Girard, ``Low-complexity quantized switching controllers using approximate
  bisimulation,'' \emph{Nonlinear Analysis: Hybrid Systems}, vol.~10, pp.
  34--44, 2013.

\bibitem{pyeatt2001decision}
L.~D. Pyeatt, A.~E. Howe \emph{et~al.}, ``Decision tree function approximation
  in reinforcement learning,'' in \emph{Proceedings of the third international
  symposium on adaptive systems: evolutionary computation and probabilistic
  graphical models}, vol.~2, no. 1/2.\hskip 1em plus 0.5em minus 0.4em\relax
  Cuba, 2001, pp. 70--77.

\bibitem{DBLP:journals/corr/abs-1810-04240}
K.~D. Julian, M.~J. Kochenderfer, and M.~P. Owen, ``Deep neural network
  compression for aircraft collision avoidance systems,'' \emph{CoRR}, vol.
  abs/1810.04240, 2018.

\bibitem{scikit-learn}
F.~Pedregosa, G.~Varoquaux, A.~Gramfort, V.~Michel, B.~Thirion, O.~Grisel,
  M.~Blondel, P.~Prettenhofer, R.~Weiss, V.~Dubourg, J.~Vanderplas, A.~Passos,
  D.~Cournapeau, M.~Brucher, M.~Perrot, and E.~Duchesnay, ``Scikit-learn:
  Machine learning in {P}ython,'' \emph{Journal of Machine Learning Research},
  vol.~12, pp. 2825--2830, 2011.

\bibitem{reissig2016feedback}
G.~Reissig, A.~Weber, and M.~Rungger, ``Feedback refinement relations for the
  synthesis of symbolic controllers,'' \emph{IEEE Transactions on Automatic
  Control}, vol.~62, no.~4, pp. 1781--1796, 2016.

\bibitem{euler}
K.~G. Larsen, A.~L. Co{\"{e}}nt, M.~Mikucionis, and J.~H. Taankvist,
  ``Guaranteed control synthesis for continuous systems in uppaal tiga,'' in
  \emph{Cyber Physical Systems. Model-Based Design - 8th International
  Workshop, CyPhy 2018, and 14th International Workshop, {WESE} 2018, Turin,
  Italy, October 4-5, 2018, Revised Selected Papers}, ser. Lecture Notes in
  Computer Science, R.~D. Chamberlain, W.~Taha, and M.~T{\"{o}}rngren, Eds.,
  vol. 11615.\hskip 1em plus 0.5em minus 0.4em\relax Springer, 2018, pp.
  113--133.

\bibitem{Bishop}
C.~M. Bishop, \emph{Pattern recognition and machine learning, 5th Edition},
  ser. Information science and statistics.\hskip 1em plus 0.5em minus
  0.4em\relax Springer, 2007.

\bibitem{NaiveBayes}
H.~Zhang, ``The optimality of naive bayes,'' in \emph{Proceedings of the
  Seventeenth International Florida Artificial Intelligence Research Society
  Conference, Miami Beach, Florida, {USA}}, V.~Barr and Z.~Markov, Eds.\hskip
  1em plus 0.5em minus 0.4em\relax {AAAI} Press, 2004, pp. 562--567.

\bibitem{jagtap2018software}
P.~Jagtap, F.~Abdi, M.~Rungger, M.~Zamani, and M.~Caccamo, ``Software fault
  tolerance for cyber-physical systems via full system restart,'' \emph{arXiv
  preprint arXiv:1812.03546}, 2018.

\bibitem{swikir2019compositional}
A.~Swikir and M.~Zamani, ``Compositional synthesis of symbolic models for
  networks of switched systems,'' \emph{IEEE Control Systems Letters}, vol.~3,
  no.~4, pp. 1056--1061, 2019.

\bibitem{rungger2015state}
M.~Rungger, A.~Weber, and G.~Reissig, ``State space grids for low complexity
  abstractions,'' in \emph{2015 54th IEEE Conference on Decision and Control
  (CDC)}.\hskip 1em plus 0.5em minus 0.4em\relax IEEE, 2015, pp. 6139--6146.

\bibitem{tomar2017invariance}
M.~S. Tomar, M.~Rungger, and M.~Zamani, ``Invariance feedback entropy of
  uncertain control systems,'' \emph{arXiv preprint arXiv:1706.05242}, 2017.

\bibitem{NairFagniniZampieriEvans07}
G.~N. Nair, F.~Fagnani, S.~Zampieri, and R.~J. Evans, ``Feedback control under
  data rate constraints: An overview,'' \emph{Proc. of the IEEE}, vol.~95,
  no.~1, pp. 108--137, 2007.

\end{thebibliography}

\clearpage

\appendix

\begin{table*}[h]
    \caption{Result of running the various methods on 10 different case studies. The `Lookup table' column gives the size of the domain of the original controller. 
    The columns `CART', `LogReg' and `MaxFreq' report the number of nodes of the decision trees constructed with the respective algorithm. 
    The column `RandomDet' reports the size of the DT that was generated by CART after the data was randomly determinized.
    The column BDD reports the size of BDDs representing the controller.}
    \label{tab:app}
    \begin{tabular}{lrrrrrr}
    	\toprule
    	Case Study & Lookup table & CART & LogReg & MaxFreq & RandomDet & BDD\\
    	\midrule
    	\multicolumn{3}{l}{\textbf{Single-input non-deterministic}}                                            \\
    	cartpole \cite{jagtap2018software}   &                                         
    	    271    & 253 & 199 & 11 & 531 & 409             \\
    	2D Thermal \cite{girard2013low}   &                                            
    	    40,311    & 27 & 27 & 9 & 27,527 & 269             \\
    	helicopter \cite{jagtap2018software}  &                                         
    	    280,539    & 6,347 & 3,753 & 229 & 454,587 & 2,313             \\
    	cruise \cite{cruise:LarsenMT15}    &                                           
    	    295,615    & 987 & 783 & 3 & 382,737 & 1,815             \\
    	dcdc \cite{SCOTS:RunggerZ16}       &                                          
    	    593,089    & 271 & 139 & 9 & 325,555 & $\infty$             \\
    	\multicolumn{3}{l}{\textbf{Multi-input non-deterministic}}                                           \\
    	10D Thermal \cite{jagtap2017quest}  &                                        
    	    26,244    & 17,297 & 147 & 7 & 42,155 & 1,012             \\
    	truck\_trailer\cite{khaled2019pfaces}      &   
    	    1,386,211    & 338,389 & $\infty$ & 43,195 & $\infty$ & 10,574             \\
    	traffic\cite{swikir2019compositional}       &      
    	    16,639,662    & 12,573 & 8,953 & 195 & $\infty$ & $\infty  $           \\
    	\multicolumn{3}{l}{\textbf{Multi-input deterministic}}                                                   \\
    	vehicle \cite{SCOTS:RunggerZ16}            &    
    	    48,018    & 13,229 & 10,375 & n/a & n/a & 7,985             \\
    	aircraft \cite{rungger2015state}        &    
    	    2,135,056    & 913,857 & 815,045 & n/a & n/a & 769,098             \\

    	\bottomrule
    \end{tabular}
\end{table*}

\section{Output of \tool\ for DT in Figure \ref{fig:10rooms}}\label{app:output}
\begin{figure}[h]
\includegraphics[width=0.45\textwidth]{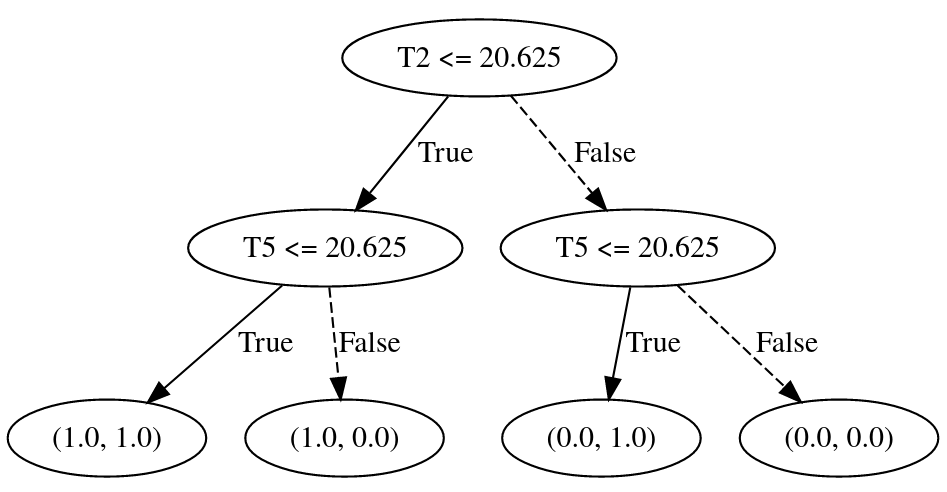}
\caption{The DOT output of \tool\ for the DT in Figure~\ref{fig:10rooms}, as displayed by Graphviz.}
\label{fig:dot}
\end{figure}

The following is the \texttt{C}-code for the DT in Figure \ref{fig:10rooms}, and Figure \ref{fig:dot} shows the corresponding DOT output.

\lstset{language=C}

\begin{lstlisting}[frame=single]  %

if (x[1] <= 20.625) {
	if (x[4] <= 20.625) {
		result[0] = 1.0f;
		result[1] = 1.0f;
	}
	else {
		result[0] = 1.0f;
		result[1] = 0.0f;
	}

}
else {
	if (x[4] <= 20.625) {
		result[0] = 0.0f;
		result[1] = 1.0f;
	}
	else {
		result[0] = 0.0f;
		result[1] = 0.0f;
	}
}
\end{lstlisting}

\section{Additional experimental results}\label{app:exp}

In Table \ref{tab:app}, we compare our algorithms as described in Section \ref{sec:exp} to the size of BDDs representing the controllers and to the idea of randomly determinizing the controller before applying the DT algorithms.
Unlike in Table \ref{tab:experiments}, we report the full number of nodes, not the number of decision paths, to make the comparison to BDDs fairer.
For clarity, we did not include all the algorithms from Table \ref{tab:experiments}. 
However, if needed, one can compute the number of nodes for every algorithm by multiplying the number of decision paths in Table \ref{tab:experiments} with two and then subtracting one.  
\end{document}